\documentclass[conference]{IEEEtran}
\IEEEoverridecommandlockouts
\usepackage{cite}
\usepackage{amsmath,amssymb,amsfonts}
\usepackage{algorithmic}
\usepackage{graphicx}
\usepackage{textcomp}
\usepackage{xcolor}
\def\BibTeX{{\rm B\kern-.05em{\sc i\kern-.025em b}\kern-.08em
    T\kern-.1667em\lower.7ex\hbox{E}\kern-.125emX}}

\usepackage{fancyhdr}
\begin{document}

\title{Investigating Large Language Models' Perception of Emotion Using Appraisal Theory}

\author{\IEEEauthorblockN{Nutchanon Yongsatianchot}
\IEEEauthorblockA{\textit{Khoury College of Computer Science} \\
\textit{Northeastern University}\\
Massachusetts, USA \\
nutjung.nutlc@gmail.com}
\and
\IEEEauthorblockN{Parisa Ghanad Torshizi}
\IEEEauthorblockA{\textit{Khoury College of Computer Science} \\
\textit{Northeastern University}\\
Massachusetts, USA \\
ghanadparisa@gmail.com}
\and
\IEEEauthorblockN{Stacy Marsella}
\IEEEauthorblockA{\textit{Khoury College of Computer Science} \\
\textit{Northeastern University}\\
Massachusetts, USA \\
s.marsella@northeastern.edu}
}

\maketitle
\thispagestyle{fancy}

\begin{abstract}
Large Language Models (LLM) like ChatGPT have significantly advanced in recent years and are now being used by the general public. As more people interact with these systems, improving our understanding of these black box models is crucial, especially regarding their understanding of human psychological aspects. In this work, we investigate their emotion perception through the lens of appraisal and coping theory using the Stress and Coping Process Questionaire (SCPQ). SCPQ is a validated clinical instrument consisting of multiple stories that evolve over time and differ in key appraisal variables such as controllability and changeability. We applied SCPQ to three recent LLMs from OpenAI, davinci-003, ChatGPT, and GPT-4 and compared the results with predictions from the appraisal theory and human data. The results show that LLMs' responses are similar to humans in terms of dynamics of appraisal and coping, but their responses did not differ along key appraisal dimensions as predicted by the theory and data. The magnitude of their responses is also quite different from humans in several variables. We also found that GPTs can be quite sensitive to instruction and how questions are asked. This work adds to the growing literature evaluating the psychological aspects of LLMs and helps enrich our understanding of the current models.
\end{abstract}

\begin{IEEEkeywords}
Large language model, Appraisal theory, coping
\end{IEEEkeywords}

\section{Introduction}

Large language models (LLM) have made significant progress in recent years. With the introduction of ChatGPT by OpenAI, the general public, not just researchers, has widely used and interacted with these LLMs. These models can write stories, songs, poems, and code. People have also used them to answer various questions, including basic facts about the world, medical questions, and social and emotional events. As these AI systems interact with people more and more, it is essential to investigate and improve our understanding of how they perceive and understand humans' social and psychological aspects. Existing research has begun to study various cognitive and psychological abilities of LLMs, including decision-making, information search, causal reasoning, and theory of mind\cite{binz2023using, bubeck2023sparks, kosinski2023theory}. 

Continuing this line of research, in this work, we aim to further investigate LLMs' ability to perceive and evaluate emotions and related factors. Emotion has multiple dimensions, including the expression of emotion, the relation to cognition, physiological experience, subjective experience, and the impact on coping responses. There are also multiple theories of emotion \cite{lazarus1991emotion, moors2013appraisal, ekman1999basic, damasio1996somatic, russell1980circumplex, barrett2017theory}. We choose to investigate emotion perception through the lens of appraisal and coping theory. Specifically, we compare LLMs perception of emotional and stressful scenarios to the characterizations of these scenarios by appraisal theory and related human data. From another angle, we investigate whether or not LLMs are sensitive to appraisal dimensions of scenarios and whether this would lead to responses with different coping tendencies. We choose appraisal theory because it provides a representation of emotional scenarios in terms of appraisal variables, allowing us to investigate emotion perception at a deeper level beyond simple emotion categories. In addition, some appraisal theories, such as Lazarus’s theory \cite{lazarus1991emotion}, provide a link from appraisal variables to coping behaviors, allowing us to further examine LLMs' responses at the behavior level. 

To accomplish this, we use a validated clinical instrument, the Stress and Coping Process Questionaire (SCPQ), by Perrez and Reicherts \cite{perrez1992stress}. SCPQ is built upon Lazarus’s appraisal and coping theory. It includes measurements of emotional experience, appraisal variables, and coping intentions and behaviors. It has also been used to evaluate a computational model of emotion before \cite{gratch2005evaluating}. In SCPQ, subjects are presented with hypothetical stereotypical stressful scenarios which evolve over time, and their responses are measured across multiple time steps. This allows us to investigate the dynamics of appraisal and coping. Furthermore, SCPQ consists of two specific types of scenarios: aversive and loss or failure. These two types differ significantly along several key appraisal dimensions: controllability, changeability, and ambiguity. This permits us to check the model’s sensitivity to appraisal dimensions. In sum, SCPQ provides a useful testbed to investigate the important aspects of appraisal and coping theory within LLMs.

We subjected SCPQ to three recent LLMs from OpenAI: \texttt{text-davinci-003}, ChatGPT, and GPT-4 \cite{brown2020language, openai2023technical}. We focus on models from OpenAI because they are the most well-known models and GPT-4 seems to be the current best available model at the time of this writing \cite{peng2023instruction}. 
We compared their results with human data and hypotheses from the theory \cite{perrez1992stress}. In addition, we tested how LLMs would change if we instructed them to act as a person with depression compared to what the theory predicted. Lastly, we also investigated the sensitivity of these models on instruction and prompts along several aspects. The results show that LLMs' responses are similar to human trends regarding the dynamics of appraisal and coping. However, they still could not differentiate between the two scenario types well. Their responses are also quite different from humans in terms of magnitude in several key variables, including controllability and coping. ChatGPT and GPT-4, when instructed to act as a depressed person, respond in a way that is consistent with the theory’s prediction. Lastly, we found that LLMs can be quite sensitive to instruction and how questions are asked.

\section{Related Work}

As SCPQ is heavily influenced by Lazarus’ appraisal and coping theory, we first briefly review Lazarus's theory here. Appraisal theories of emotion define appraisal as an evaluation of what the situation implies for personal well-being based on one's goals and beliefs \cite{arnold1960emotion}, \cite{smith1990emotion}, \cite{lazarus1991emotion}, \cite{moors2013appraisal}. Lazarus's theory emphasizes the importance of the process or dynamics involved in coping \cite{lazarus1991emotion}. In particular, the person-environment relationship is always changing, leading to different, evolving emotional experiences, appraisal evaluations, and coping.

Lazarus proposes two main dimensions of appraisals: primary and secondary appraisal dimensions. Primary appraisals include goal relevance, goal congruence, and type of ego-involvement. Secondary appraisals include blameworthiness, coping potential (whether and how a person can manage the demands and consequences of the situation), and future expectancy (the degree to which things are likely to change for the better or worse ). Effectively, secondary appraisals involve how people can cope with the situation. Note that, in SCPQ, with influence from earlier work on helplessness \cite{seligman1972p}, Perrez and Reicherts use the term controllability (the subjective appraisal of personal ability to control the situation) instead of coping potential and changeability (the subjective appraisal that the stressful event will change by itself) instead of future expectancy.  

Lazarus also proposes two broad types of coping: problem-focused coping (directly changing the situation or the environment) and emotion-focused coping (changing one's goals and/or beliefs to adjust to the situation). These copings are also the main focus of SCPQ. 

With the influence of Lazarus's theory, SCPQ focuses on not only appraisal but also the dynamics of appraisal and coping. This makes it stand out among other similar scenario-based instruments \cite{harmon2016discrete, scherer2021evidence}. In addition, SCPQ extends Lazarus's taxonomy further. We go into more detail in the next section. Additionally, SCQP has been used to evaluate a computational model before \cite{gratch2005evaluating}. A critical difference is that in the previous work, the scenarios were manually constructed to be in the right format that the model could process, but here we are using LLMs to interpret the scenario directly from the text.

On the other side, there has been more and more work evaluating the psychological aspects of LLMs. For example, Binz and Schulz (2023) studied GPT-3's decision-making, information search, and causal reasoning using cognitive psychological tests such as heuristic and biases tests and the cognitive reflection tests \cite{binz2023using}. They found that it can solve these tasks similarly or better than human subjects. Kosinski (2023) investigated Theory of Mind (ToM) in LLMs using standard false-belief tasks and found that ChatGPT and \texttt{text-davinci-003} can solve most ToM tasks \cite{kosinski2023theory}. Miotto et al. (2022) explored personality, values, and demographic of GPT-3 using validated questionnaires \cite{miotto2022gpt}. They found GPT-3 to be similar to the human baseline sample and is close to a young adult demographic. Bubeck et al. (2023) subject GPT-4 to various tests such as mathematics, coding, medicine, law, and psychology \cite{bubeck2023sparks}. They show that GPT-4 outperforms ChatGPT on ToM and emotion perception. Nevertheless, they simply tested the models on a few examples and did not systematically evaluate their psychological aspects and related factors.      

\section{Stress and Coping Process Questionaire}

The Stress and Coping Process Questionaire (SCPQ) was developed by Perrez and Reicherts to measure a human subject’s appraisal and coping variables in stressful and emotional scenarios that occur in their daily life \cite{perrez1992stress}. SCPQ has been validated by a panel of clinician experts and applied to normal human subjects as well as in clinical settings. 

A subject is presented with a series of hypothetical scenarios that are divided into three episodes or phases, corresponding to different stages of the stressful scenario: phase 1 beginning, phase 2 continuation, and phase 3 outcome. Their responses are measured at the end of each phase, reflecting the key assumption of SCPQ that the dynamics of a stressful scenario are crucial to understanding how stress and coping develop.   

SCPQ consists of two types of scenarios: aversive and loss or failure (loss). 
Examples of loss scenarios are the loss of a friendly relationship, the loss of an important object, and the failure of an interesting side job. Examples of aversive scenarios are criticism from the partner, arguments about problems in a relationship, and reproaches from colleagues. The key differences between the two types are the level of controllability, changeability, and ambiguity. By design, \textbf{the loss scenarios are less controllable, less changeable, and less ambiguous than the aversive scenarios.}

Both types of scenarios follow a similar course of three episodes. The loss or aversive scenario is looming at the beginning (phase 1) and \textbf{becomes unavoidable, imminent, or reinforced in phase 2}. The outcome phase (phase 3) can either be positive or negative. For loss scenarios, the positive outcome involves finding a substitution, while the negative outcome depicts the final loss without any successful substitution. For aversive scenarios, the positive outcome involves successfully removing the source of stress, while the negative outcome depicts the continuation of the stress. 

Below are examples of an aversive scenario and a loss scenario, respectively. 

An aversive scenario with a positive outcome:
\begin{itemize}
    \item Phase 1: "You are together with some colleagues. One says that you don't pull your weight when there is difficult work. He claims that you don't think of other colleagues."
    \item Phase 2: "Sometime later, another colleague hints that the problem is not that you don’t think of others but that you lack any real interest in the work."
    \item Phase 3: "Finally, you realize what your colleagues were really getting at, and you, for your part, were able to convince them that you sometimes are more cautious at your work than others."
\end{itemize}

A loss scenario with a negative outcome.
\begin{itemize}
    \item Phase 1: "A person who was very close to you, especially in recent times, has to move away unexpectedly. When you parted, you reassured each other you would both keep in close contact. But his/her new home is quite far away. You could see each other only rarely, if at all."
    \item Phase 2: "In the meantime, some weeks have passed. The person hasn’t gotten in touch with you again. Nevertheless, you feel from time to time that you miss him/her."
    \item Phase 3: "Finally, it has become clear that your friendship is not the same anymore. Your relationship with other people can’t replace what you have lost. Now and then, you feel disappointed about the relationship you have lost."
\end{itemize}

There are nine scenarios for each type, a total of eighteen scenarios. The responses can be aggregated to reflect the general tendency toward these types of scenarios and compared between the two types, which differ along crucial appraisal dimensions.

SCPQ includes the following measurement.

\begin{itemize}
    \item Emotional Responses: 1) anxious - calm, 2) depressed - cheerful, and 3) angry - gentle,
    \item Appraisals: 1) changeability, 2) controllability, and 3) negative valence,
    \item Coping intentions: 1) Problem-focused coping, 2) Emotion-focused coping\footnote{The question is “To remain calm and composed …” Strictly speaking, this is not the same as emotion-focused coping as defined in Lazarus theory which is about changing one internal beliefs, goals, or intention.}, and 3) Self-esteem, 
    \item Self-directed coping behaviors: 1) search for information, 2) suppress information, 3) re-evaluation, and 4) palliation (calming self-instruction or smoking, drinking, and eating),
    \item Environment-directed coping behavior: 1) Active (to prevent or confront the stressor) and 2) Passive (waiting, hesitating, resigning). 
    \item Blameworthines: 1) Self-blaming and 2) Other-blaming,
\end{itemize}

Below, we summarize the hypotheses that are supported by the human data from the SCPQ study\footnote{Note that we do not present the results involving self-directed coping here as they were not supported by human data, but the LLM results can be found on Github.}.

\begin{itemize}
    \item H1.1: Valence should be lower in the positive outcome than in the negative outcome in phase 3.
    \item H1.2: Subjects should perceive higher controllability and changeability in the aversive scenarios than in the loss scenarios.
    \item H1.3: Controllability and changeability should decrease from phase 1 to phase 2.
    \item H2.1: Subjects should use more active coping in aversive scenarios than in loss scenarios.
    \item H2.2: Subjects should use less passive coping in aversive scenarios than in loss scenarios.
    \item H3.1: Subjects’ intention to use problem-focused coping is less in aversive scenarios than in loss scenarios.
    \item H3.2: Subjects’ intention to use emotion-focused coping is more in aversive scenarios than loss scenarios.
    \item H4.1: Subjects will blame themselves and others more in aversive scenarios than in loss scenarios.
    \item H4.2: Self-blame will decrease over time, while Other-blame will increase over time.
\end{itemize}

These are the trends that we will investigate in LLMs’ results. The main rationale of H2-H4 is that aversive scenarios should be perceived as more controllable and changeable, so subjects are expected to cope differently between the two types of scenarios. The SCPQ study involved 100 non-student adults with an average age of 38 years (sd 11.8). 

Additionally, Perrez and Reicherts provide the following hypotheses regarding depression: 

\begin{itemize}
    \item H5.1: Depressed persons perceive stressful scenarios to be more stressful and higher negative valence.
    \item H5.2: Depressed persons perceive lower controllability and changeability. 
    \item H6.1: Depressed persons use less active/problem-focused coping.
    \item H6.2: Depressed persons use more palliation.
    \item H6.3: Depressed persons blame themselves more. 
\end{itemize}

In short, depressed persons are expected to perceive scenarios worse both in controllability and changeability, resulting in different coping patterns.

\section{OpenAI's GPTs}

In this work, we choose to investigate three recent LLMs from OpenAI's family of Generative Pre-trained Transformer models, or GPT \cite{brown2020language, openai2023technical}. These include 
\texttt{text-davinci-003} (D003), \texttt{gpt-3.5-turbo} (ChatGPT), \texttt{gpt-4} (GPT-4). The first two are from the GPT3.5 family. These three models have been fine-tuned using Reinforcement Learning with Human Feedback (RLHF) \cite{ouyang2022training}, and ChatGPT and GPT-4 have been optimized for chat. ChatGPT and GPT-4 also allow the user to set a system message (i.e., describing what kind of an assistant you want it to be). We do not use this feature to allow a comparison with the old model. To maximize the replicability of our results, we set the temperature parameter to 0 in all of our experiments. This makes the outputs mostly deterministic, selecting the outputs with the highest log probability. All other parameters are set to default. 

As these models can be sensitive to instruction \cite{bommarito2022gpt, li2022gpt, binz2023using}, we investigate four different variations of prompting and asking the models. 
Here is the default instruction taken from SCPQ with a slight modification: “Try to clearly imagine the scenario below and then answer the question with the choice only in one line.”
First, we either ask it to output choices (default) or just the number only (“the choice’s number only”). The number only makes sense here because all measurements use a Likert scale ranging from 0 up to 5. We test this variation because our early testing showed that sometimes the models may output more than just a choice, such as repeating the question, even when the instruction specifies “choice only."   

The second variation is the location of the instruction. There are two versions: either putting the instruction before (default) or after (“the above scenario”) the scenario. The reason for testing this is that, as these models use attention mechanisms, the distance of the context could impact how the LLM follows the instruction.

Third, we investigate either asking them one question at a time (individual) or multiple questions at a time (batch). The batch follows the set of questions as stated above. The rationale for this is that asking in batches can save time and costs, as you don’t need to repeat the scenario every time.

These first three variations result in eight different combinations of instructions. Lastly, we also test the effect of appending the previous (appraisal) answers to the prompt. The reason is that, as we are interested in the dynamics, knowing their previous answers could be crucial. For this variation, we only use the default instruction as asking for the number only or after the scenario does not make sense in this case. 

Code, including all SCPQ scenarios and instructions, data, and all results, including additional results not shown in the paper, can be found at github.com/yongsa-nut/PerrezSAIWS.

\section{Results}

\begin{figure*}[htbp]
\centering
\includegraphics[scale=0.9]{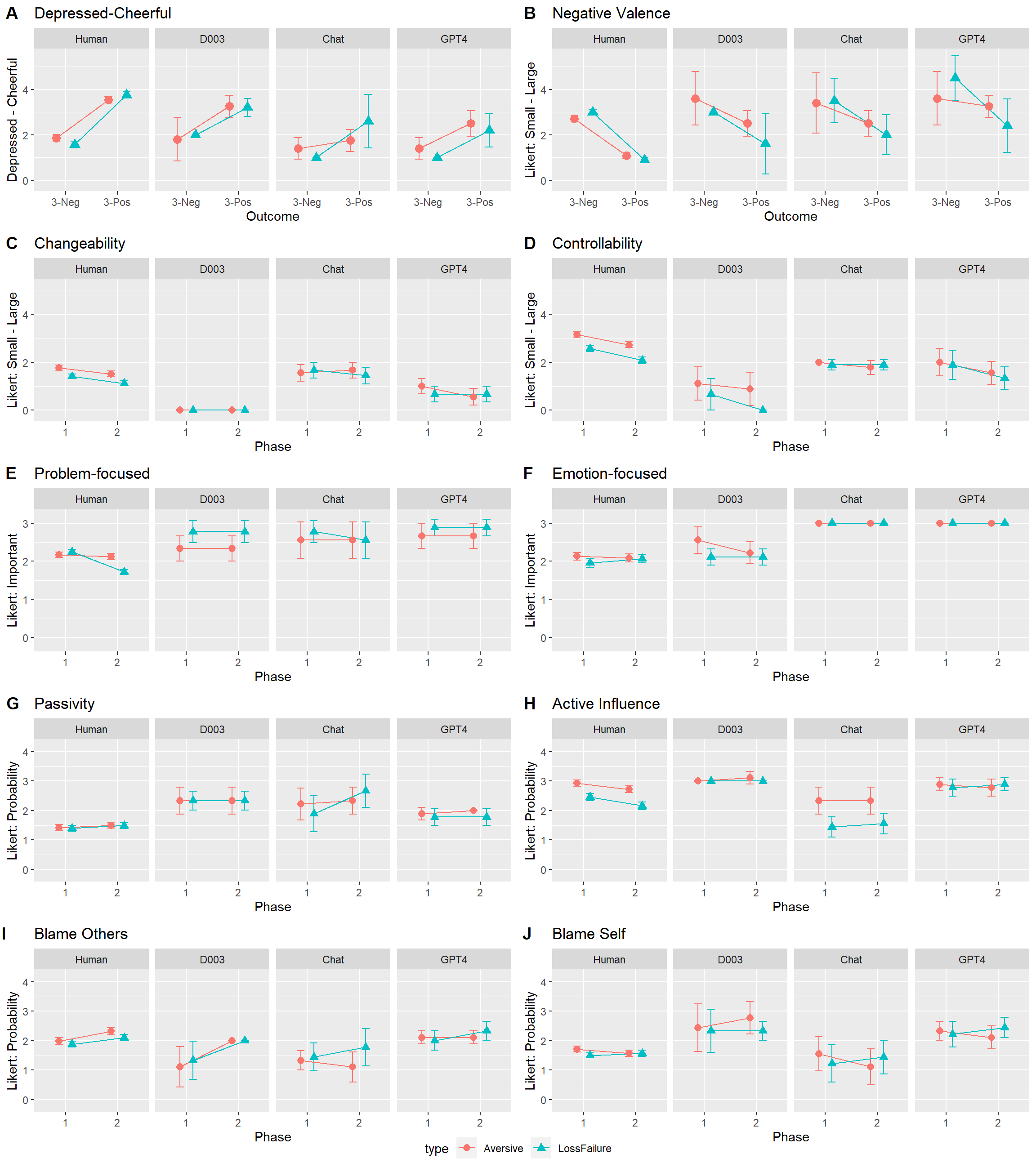}
\caption{Human vs The three models results for selected variables. The points show The estimated means and the error bars is 95\% standard errors. The pink line with circle dots is the aversive type and the blue line with triangles is the loss type. The likert scales are as follows. Emotion: Very depressed (0) - Very cheerful (5); Appraisal: Very small (0) - very large (5); Coping Intention: Not important (0) - Very important (4); and Coping behaviors: Not at all 0\% (0) - Certainty 100\% (4).}
\label{fig:main_results}
\end{figure*}

Figure \ref{fig:main_results} shows the estimated mean with the 95\% standard error for all the key measurements of the three models and human data. The setup here is the default setup, where the question is asked one by one, and the instruction is placed before the scenario and asks for choices. We choose to report this here as it is the most similar to the human setup. We discuss the results for other setups in the next section. 

Crucially, we focus mainly here on the qualitative results comparing the trend of results from the model and humans. The main reason is that there is a discrepancy between human data and model data. The human results are obtained from averaging over 100 subjects and nine scenarios, while the model results are from averaging nine scenarios making their uncertainty incomparable. 

Figure \ref{fig:main_results}.A shows the results for depressed/cheerful emotional reactions. For this result and valence, we only focus on the outcome (positive or negative) in phase 3. We see that all three models show the expected trend where the positive outcome results in more cheerful and less depressed than the negative outcome. Compared to humans, all three models rate the cheerful to be lower in the positive outcome, where D003 is closest to the human rating. The results for the other two emotional reactions are similar.

The results for valence in Figure \ref{fig:main_results}.B also shows a similar trend. Like humans, all three models rate the valence to be lower in the positive outcome than in the negative outcome. However, all three models rate valence higher than humans in both negative and positive outcomes. 

Next, for changeability in Figure \ref{fig:main_results}.C, we see that none of the models follow the human trend exactly where there is a difference between the two types of scenarios across two times, and the changeability in both types goes down. D003 always rates changeability to be zero. On the other hand, ChatGPT only rates changeability to go down in phase 2 for loss scenarios, while GPT-4 only rates changeability to go down for aversive scenarios. For controllability (Figure \ref{fig:main_results}.D), we see that only D003 and GPT-4 show the expected trend of controllability going down over time for both scenario types. However, GPT-4 does not perceive the two types to be different, unlike D003. In all cases, all three models perceive controllability to be lower than what humans perceive.

We turn now to coping intentions. For problem-focused coping in Figure \ref{fig:main_results}.E, only ChatGPT shows the trend of lowering it over time for loss scenarios. None of the models show that problem-focused coping at phase 2 in loss scenarios is lower than in aversive scenarios. In addition, all models rate problem-focused coping higher than the human data across time and type. For emotion-focused coping in Figure \ref{fig:main_results}.F, we see that only D003 shows a similar trend to the human data, where the intention is going down over time in the aversive case. On the other hand, both ChatGPT and GPT-4 rate it maximum across time and type. 

Next, we look at coping behaviors. First, for passivity (Figure \ref{fig:main_results}.G, both ChatGPT and GPT-4 show a trend similar to humans where the passivity increases over time.
Second, for active influence (Figure \ref{fig:main_results}.H), only GPT-4 shows the trend that the active influence would decrease over time but only for the aversive case. On the other hand, only ChatGPT shows a clear difference between the two types. 

Lastly, we turn to blameworthiness. First, for blaming others (Figure \ref{fig:main_results}.I), all models show that, in the loss scenarios, blaming others increases from phase 1 to 2. However, only D003 shows an increase in blaming others in the aversive scenarios. None of the models shows that blaming others is higher in the aversive than in the loss scenarios at phase 2, like the human data.

Second, for self-blaming (Figure \ref{fig:main_results.}J), both ChatGPT and GPT-4 show trends similar to the human data, where blaming oneself decreases over time in the aversive type and is higher in the aversive type than in the loss type in phase 1.

Overall, we observe in many cases that LLMs's responses are similar to human's data in the case of the dynamics, but not in the case of scenario types. 

\begin{figure*}[htbp]
\centering
\includegraphics[scale=0.7]{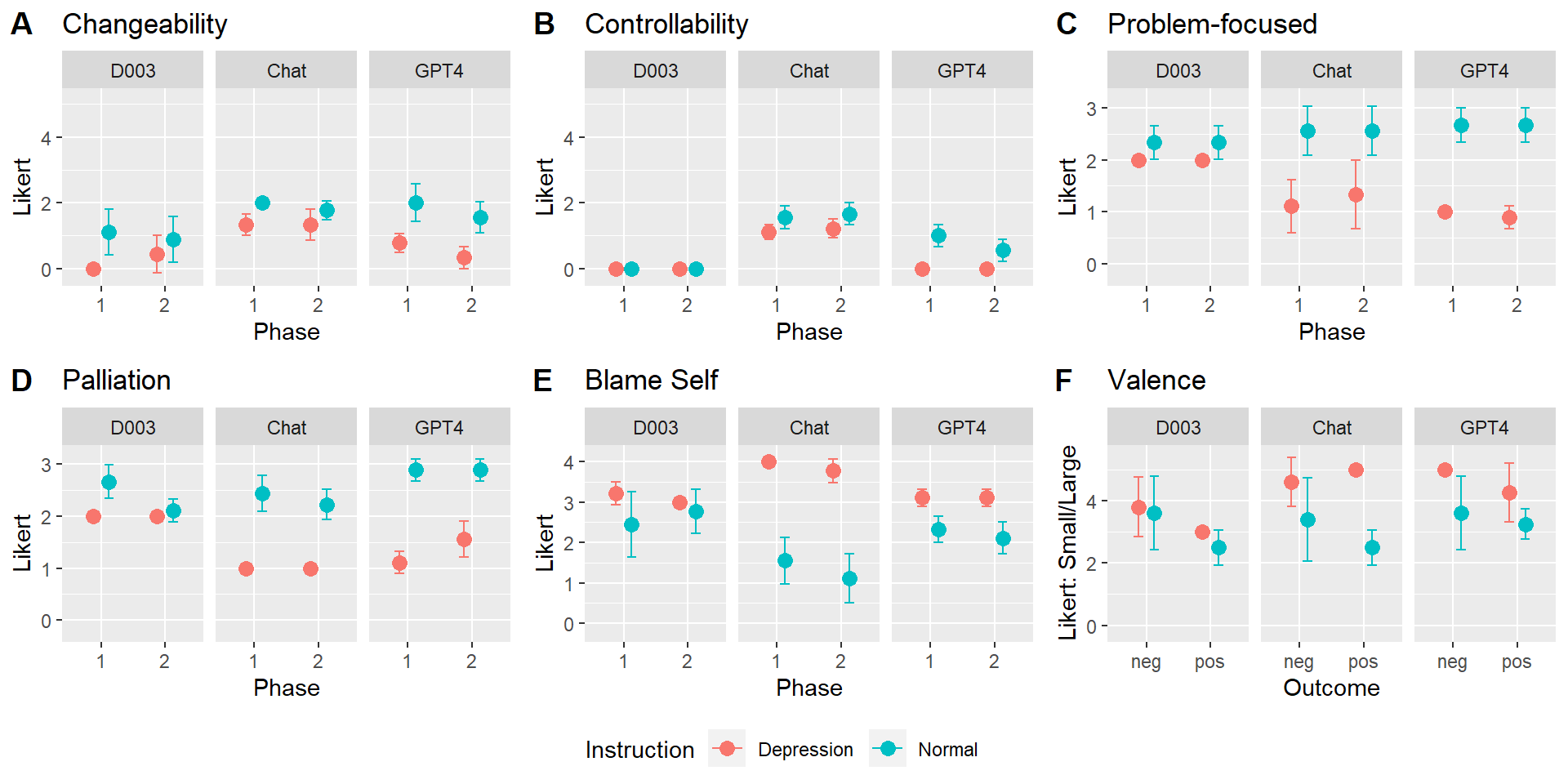}
\caption{Depression vs Normal Results for the three models for the selected variables. The pink with circle points is the depression instruction and the blue with triangle points is without the instruction. }
\label{fig:depressed_results}
\end{figure*}

Next, we look at the results comparing the model instructed to act as a person with depression (Depression) and the model without the instruction (Normal), focusing only on aversive scenarios (the loss scenarios show similar trends).  Figure \ref{fig:depressed_results} shows the key six measurements. The pattern is clear that, for ChatGPT and GPT-4 but not D003, there is a difference between the depression and normal case in the expected directions. In particular, controllability, changeability, problem-focused coping, and palliation are lower in the depression case than in the normal case, while blaming oneself and valence are higher in the depression case than in the normal case. 

\begin{figure}[htbp]
\centering
\includegraphics[scale=0.42]{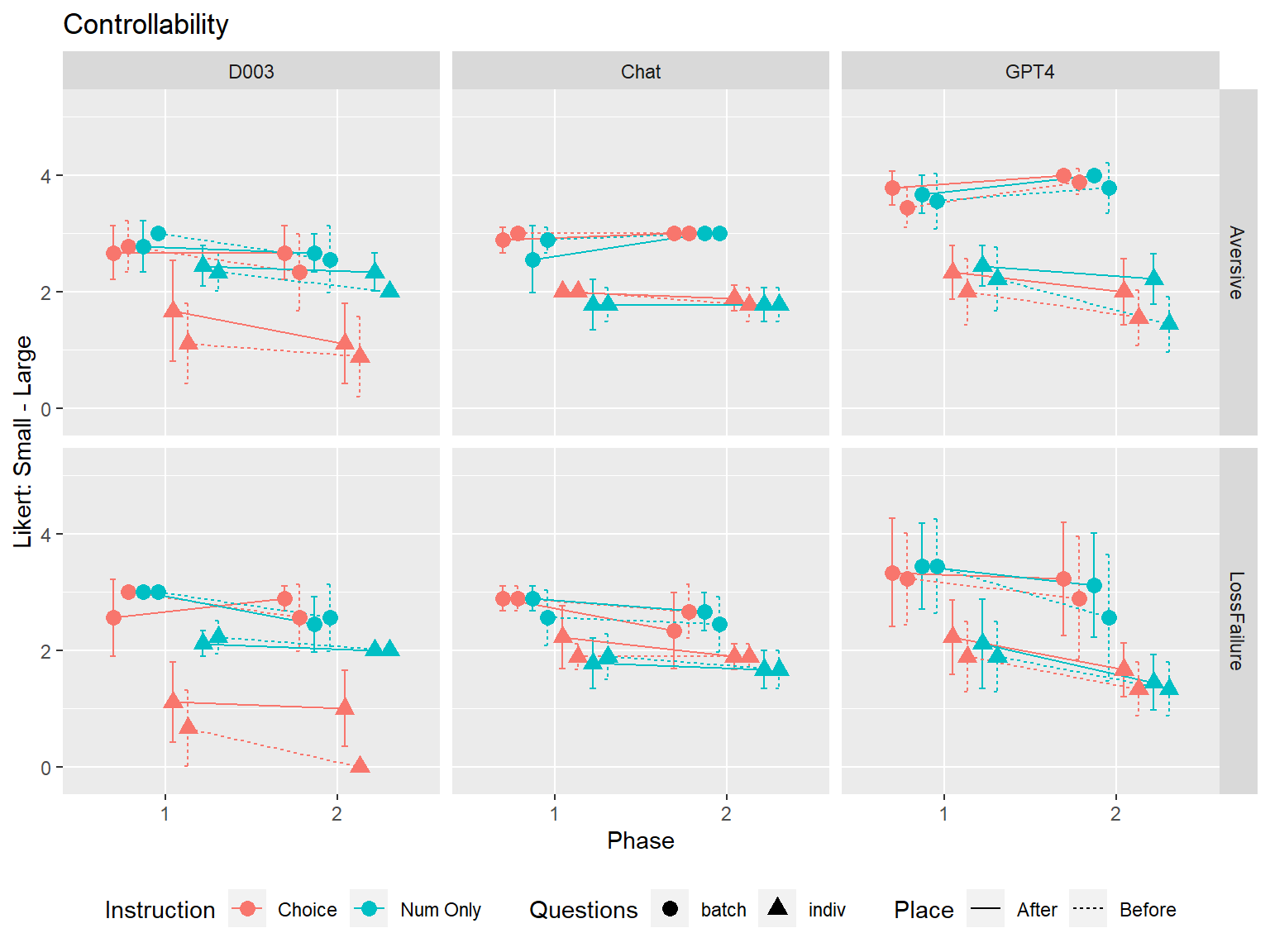}
\caption{The sensitivity analysis results on controllability for the three models across eight possible combinations across three choices. indiv = individual. Num only = Number only.
}
\label{fig:instruction_results}
\end{figure}

\begin{figure}[htbp]
\centering
\includegraphics[scale=0.5]{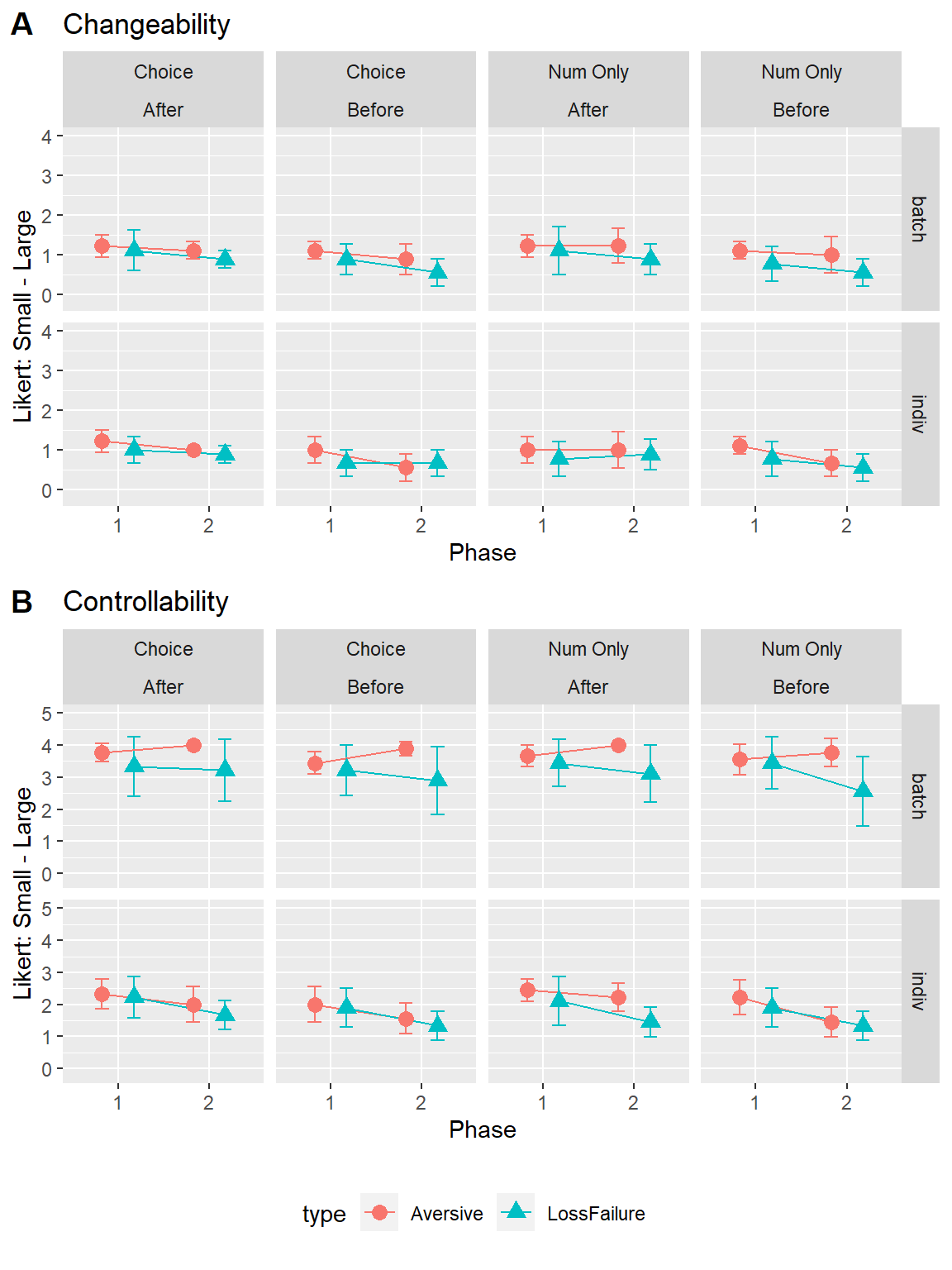}
\caption{The sensitive analysis results across eight combinations across three choices for GPT-4 on changeability (A) and controllability (B). indiv = individual. Num only = Number only.
}
\label{fig:GPT4_all_examples}
\end{figure}

Figure \ref{fig:instruction_results} shows the results on controllability for the three models across eight combination instructions across three choices. Overall, we see that there are variations across these instructions. This means that the instruction, where it is, and how many questions are asked could affect the output from the models. The biggest difference comes from asking in a batch instead of asking each question individually. The variation also depends on the model. Similar results can be found in other questions not shown here.

Next, we zoom into selected questions. Figure \ref{fig:GPT4_all_examples} shows the GPT-4’s results for changeability (A) and controllability (B) across all combinations of setup. Due to space limitations, we focus only on these two as the theory argues they strongly influence the coping response, and GPT-4 is the latest model. Again, we see that there are variations in both controllability and changeability across combinations. For changeability (Figure \ref{fig:GPT4_all_examples}.A), a few combinations show the expected trends aligning with human data, where changeability decreases over time and differs between aversive and loss types. In the case of controllability (Figure \ref{fig:GPT4_all_examples}.B), it increases rather than decreases over time for the aversive type when asking in a batch. In addition, the value is also higher in the batch setup. On the other hand, when asking the questions individually, controllability decreases over time, aligning with the expected trend. However, only in one of the setups (asking to output only a number and after the scenario), controllability across all phases is higher in the aversive scenarios than in the loss scenarios, as expected by the theory and human data. Nevertheless, the value in this setup is still lower than humans, and its changeability does not align with humans. Overall, there is no single setup here where both changeability and controllability align with the expected trends.

In addition to these eight setups, we look at the effect of appending their appraisal answers to the prompt. However, we do not observe any significant changes in any variables aside from a few cases for ChatGPT. These include changeability and controllability in phase 2, in the right direction.    

Beyond the variation shown in the figure, we found that GPT-4 follows instructions better than the other two models. In particular, when asking in a batch, ChatGPT and D003 may not answer all the questions. Further, when asked to answer with choice, ChatGPT occasionally did not answer just a choice but provided a full sentence reiterating the question instead. These did not happen with GPT-4. 

\section{Discussion}

Overall, no model follows all the human trends and hypotheses as predicted by appraisal and coping theory. Nonetheless, the responses from the three models depict the right trends for the dynamics in several variables, including emotional responses, appraisal variables, and coping. In many cases, however, the models could not differentiate the two scenario types well, and the magnitudes are quite different from humans. A few cases stand out. For example, all models rate the negative valence to be more negative than humans. One potential explanation could be from the human side, namely it could be due to experimenter demand effects. Another interesting case concerns the particular aspects of emotion-focused coping that SCPQ considers, specifically to remain calm and composed. Both ChatGPT and GPT-4 always answer the highest value. We speculate that this could be due to fine-tuning with RLHF.  

Importantly, we also observe some differences between humans and LLMs on several key appraisal variables. In particular, GPT-4 rated the controllability and changeability decrease over time but didn’t rate the two scenario types differently. We speculate that this could be due to the limited information provided in the scenarios. Human subjects bring with them their own knowledge and experiences of these daily stressful scenarios, which could make them aware of various ways that they could deal with them. However, these are not explicitly in the sceanrios, and LLM may not be able to infer them from just a short snippet. Another explanation and limitation of SCPQ is that these scenarios are hypothetical, and people may behave and appraise them differently if they were real. To fully test the perception of appraisal and emotion, future work is needed to compare LLMs' results with human data from real events. 

Another interesting result is that ChatGPT and GPT-4 can be instructed to act as a depressed person, where their responses show trends similar to the theory's prediction, such as perceiving less controllability and more negative valence. Nevertheless, we need to interpret this result with caution. At a minimum, it could mean that these models have learned the stereotypical behaviors of depressed people. Future research is needed to further explore LLMs in this direction. Still, this opens up the possibility of instructing the models to act as a person with various personalities or psychological conditions to investigate how it would affect the appraisal evaluation and emotional experiences.  

This highlights another limitation of this work: human data is an average over multiple people and not a single individual. We did not compare LLMs, which have been fine-tuned in a specific way, against a specific person. Future work may look into instructing the model to match with a specific subject or group of subjects for comparison, a matched pair design. 

Our results also indicate that all models can be quite sensitive to the instruction and prompts. Asking in a batch, which could reduce the cost and speed up the query, could yield different results from asking each question one by one. Moreover, the older models may struggle to answer all the questions in the right format, especially when the number of questions increases. 

In conclusion, this work seeks to understand LLMs through the lens of appraisal and coping theory, and we found some evidence suggesting that there is still some discrepancy between how human and LLMs perceive emotional scenarios. Nevertheless, as mentioned, this only touches a few aspects of emotional experiences and provides only one view of emotion theory. It is also possible that these LLMs trained on a large amount of human data would learn a different representation of scenarios from appraisal theory. It is an open question whether or not this different representation could be used in some way to inform theory or our understanding of emotion. 

Regardless, as these black box LLMs interact with more and more people, it is crucial for researchers to investigate how they understand human emotional experiences thoroughly. This work provides some initial steps toward this endeavor.

\section*{Ethical Impact Statement}

In this work, we evaluate LLMs on their emotion perception ability. There are several ethical problems associated with LLMs including bias, harmful content, misinformation, and privacy concerns. However, given how LLMs are positioned to impact us, it is critical for research to explore and evaluate them. We did not collect human data in this work. We used existing data and results from a previously published and approved study.

\bibliographystyle{ieeetr}
\bibliography{references}

\end{document}